\documentclass[final]{cvpr}

\usepackage{times}
\usepackage{epsfig}
\usepackage{graphicx}
\usepackage{amsmath}
\usepackage{amssymb}

\usepackage{multirow} 
\usepackage{graphicx}
\usepackage{booktabs}
\usepackage{subcaption}
\usepackage{amssymb}
\usepackage{multirow,xspace}
\usepackage{makecell}
\usepackage{cite}
\usepackage{xcolor}
\usepackage{algorithm}
\usepackage{algorithmic}

\usepackage{caption}
\usepackage[pagebackref=true,breaklinks=true,colorlinks,bookmarks=false]{hyperref}

\def\cvprPaperID{6619} %
\def\confYear{CVPR 2021}

\begin{document}

\title{Few-Shot Incremental Learning with Continually Evolved Classifiers}

\author{Chi Zhang$^1$\thanks{~indicates equal contribution.}, ~~ Nan Song$^1$$^*$,~~ Guosheng Lin$^1$\thanks{Corresponding author: G. Lin (e-mail: {\tt gslin@ntu.edu.sg})}, ~~ Yun Zheng$^2$, ~~ Pan Pan$^2$, ~~ Yinghui Xu$^2$\\
	$^{1}$ Nanyang Technological University, Singapore
	~ ~ ~ 
	$^{2}$ Alibaba DAMO Academy\\
{\tt \{chi007,nan001\}@e.ntu.edu.sg}, ~
        {\tt gslin@ntu.edu.sg}
        \\  {\tt \{zhengyun.zy,panpan.pp\}@alibaba-inc.com}, ~
	{\tt renji.xyh@taobao.com}
}

\maketitle

\begin{abstract}
Few-shot class-incremental learning (FSCIL) aims to design machine learning algorithms that can continually learn new concepts from a few data points, without forgetting knowledge of old classes.
The difficulty lies in that limited data from new classes not only lead to significant overfitting issues but also exacerbate the notorious catastrophic forgetting problems. Moreover, as training data come in sequence in FSCIL, the learned classifier can only provide discriminative information in individual sessions, while FSCIL requires all classes to be involved for evaluation. In this paper, we address the FSCIL problem from two aspects. First, we adopt a simple but effective decoupled learning strategy of representations and classifiers that only the classifiers are updated in each incremental session, which avoids knowledge forgetting in the representations. By doing so, we demonstrate that a  pre-trained backbone plus a non-parametric class mean classifier can beat state-of-the-art methods. Second, to make the classifiers learned on individual sessions applicable to all classes, we propose a Continually Evolved Classifier (CEC) that employs a graph model to propagate context information between classifiers for adaptation.  To enable the learning of CEC, we design a pseudo incremental learning paradigm that episodically constructs a pseudo incremental learning task to optimize the graph parameters by sampling data from the base dataset. Experiments on three popular benchmark datasets, including CIFAR100, miniImageNet, and Caltech-USCD Birds-200-2011 (CUB200), show that our method significantly outperforms the baselines and sets new state-of-the-art results with remarkable advantages.

\end{abstract}

\section{Introduction}

Deep Convolutional Neural Networks have gained remarkable success in many computer vision tasks~\cite{resnet,liu2020weakly,li2021cyclesegnet,sun2020conditional,zhang2018efficient}, stemming from the availability of big curated datasets, along with unprecedented computing power.
However, a classification model that is trained by supervised learning can only make predictions on a set of pre-defined image categories. If we want to extend a trained model on new classes, a large amount of labeled data for new classes as well as data from old classes are both necessary for network finetuning, which inevitably hinders its real-world applications. 
If the dataset of old classes is no longer available, directly finetuning a deployed model with new classes can lead to the notorious catastrophic forgetting problem that knowledge about old classes is quickly forgotten~\cite{knowledgeDis,shin2017,kirkpatrick2017overcoming}. In contrast to machine learning systems, humans are readily able to learn a new concept with few examples without forgetting old knowledge.
The gap between humans and the machine learning algorithms fuels interest in few-shot class-incremental learning (FSCIL)~\cite{TOPIC}, which aims to design machine learning algorithms that can be continually extended to new classes with only a few data points.
The challenge of FSCIL lies in that the scarcity in the data of new classes will not only cause severe overfitting but also exacerbates the catastrophic forgetting problem of old classes.
In this paper, we undertake the task of few shot incremental learning and consider to solve the aforementioned problems from two aspects.

\begin{figure}[t]
	\centering
	\includegraphics[width=1\linewidth]{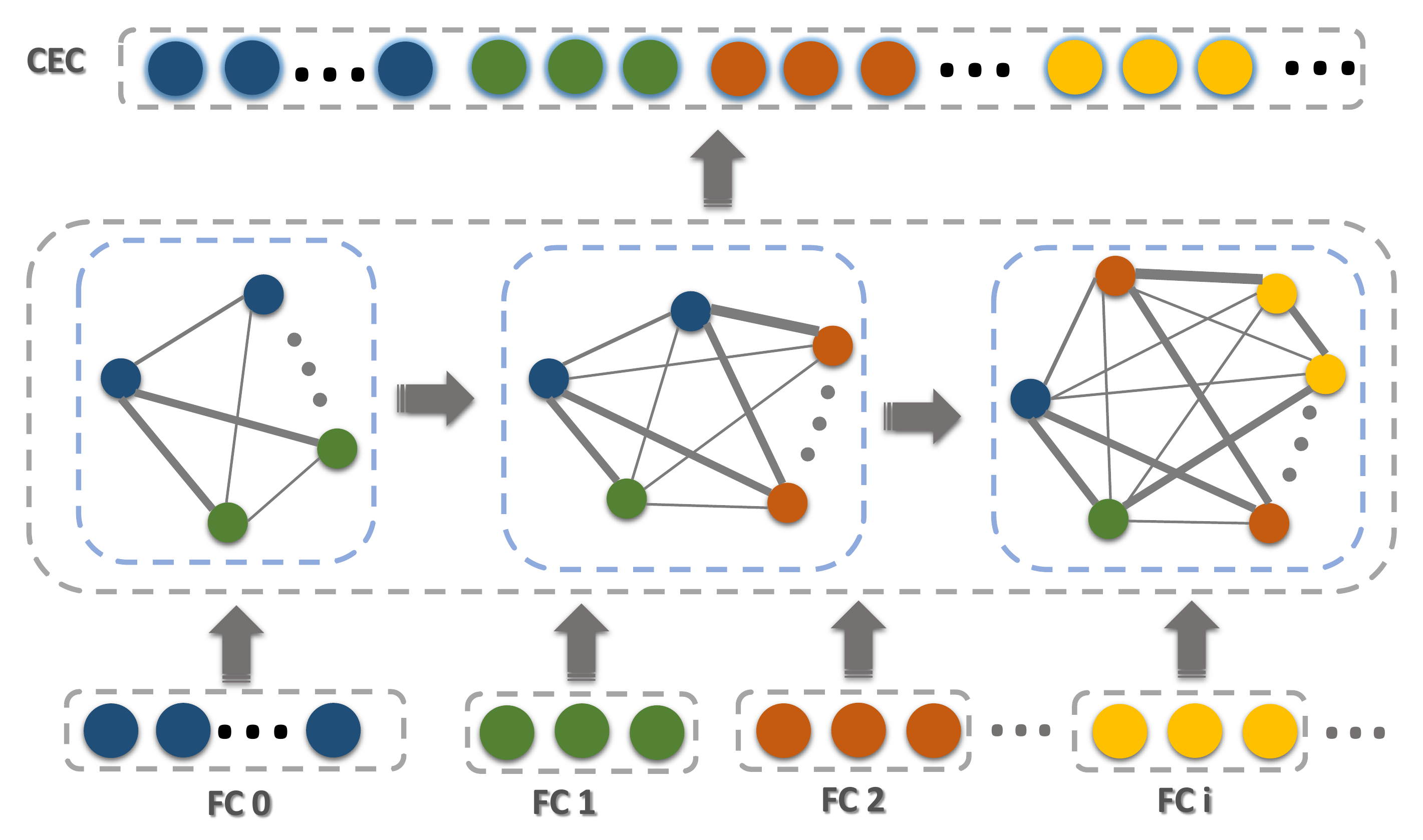}
	\caption{Illustration of our proposed continually evolved classifiers for FSCIL. We employ a graph model to adapt the classifier weights learned on individual sessions for the prediction over all classes.}
	\label{fig:small}
\vskip -1em
\end{figure}

First, as the data from base classes and new classes are severely unbalanced, 
we propose to decouple the learning of representations and classifiers for the FSCIL problem.
Specifically, the model only learns the representations in the first session where abundant data from base classes are available, and in the new sessions,  we fix the network backbone and only adapt the classifier for new classes.  Thus, we can avoid the overfitting problem  as well as the catastrophic forgetting problem in the representations.
By doing so, we demonstrate that a pre-trained network backbone based on data from base classes plus a class mean classifier can beat state-of-the-art approaches.

Second,  as the classifiers are always learned from the classes in individual incremental sessions, they can only provide discriminative information for classifying internal categories, while incremental learning aims to learn models that can apply to all classes.
As a result, even if a classifier can learn a well-separated decision boundary for the previous classes, it may lose the generalization ability when more novel classes are involved.
For example, a vehicle-related representation  \emph{wheel} is chosen by the classifier as a discriminative representation to distinguish the categories \emph{car}, \emph{dog}  and \emph{cup} in the current classification task. However, when a new category \emph{trunk} is involved in the new sessions, such representation may not be discriminative enough to classify all categories. 
Therefore, the incremental learning algorithm should have the flexibility to adjust the classifiers in previous sessions based on the overall task context to undertake the entire classification task.
To this end, we present a Continually Evolved Classifier (CEC) that can progressively adapt the classifier weights based on current and history tasks.
At the core of our network is a classifier adaptation module which uses a graph attention network (GAT)~\cite{velickovic2018graph} to 
adapt the classifier weights learned on each task.
By contextualizing individual classifier weights over the global task, the adapted classifiers highlight the discriminative representations in the backbone and generate better decision boundaries over all involved classes.

To enable the learning of the proposed continually evolved classifier, it is important to optimize the graph model under an incremental learning scenario.
However, in incremental learning, datasets from different training sessions can never be accessed simultaneously for training.
To overcome the issue, we propose a pseudo incremental learning paradigm, where we episodically construct a pseudo incremental learning task from the dataset in the base session to simulate the incremental learning scenario for training.
Our design takes inspirations from the meta-learning paradigm~\cite{matchnet}. In each pseudo incremental learning episode, we first sample a set of classes from the base dataset to play the role of the base classes, then we sample another group of classes to play the role of incremental classes to learn the model.
However, as the pre-trained backbone has already learned feature representations that can well classify the base classes, directly using the sampled classes from the base dataset for learning may bypass the GAT and thus fail to impose context knowledge. We solve this problem by randomly rotating the sampled pseudo incremental classes with a large angle to synthesize new classes.
In this way, we intentionally synthesize unfamiliar classes at training time to enforce context knowledge propagation in the graph model. Once the graph model is learned, we can use the graph model to update the classifier weights learned in incremental sessions.

To validate the effectiveness of our proposed method, we conduct comprehensive experiments on multiple benchmark datasets. The contribution of this work is summarized as follows:
\begin{itemize}
\itemsep -0.1cm 
	\item We adopt a decoupled training strategy for representation learning and classifier learning to avoid knowledge forgetting and overfitting in the backbone.
	\item We propose a continually evolved classifier that employs a graph model to combine classifiers learned on individual sessions for incremental learning.
	\item To enable the learning of the graph model in CEC, we design a pseudo incremental learning paradigm.
	\item Experiments on the CIFAR100, CUB200 and miniImagenet datasets show that our method significantly outperforms the baselines and sets new state-of-the-art performance with remarkable advantages.
\end{itemize}

\section{Related Work}
\textbf{Few-Shot Learning.} Few-shot learning aims to learn a model that can classify unseen images when only training from scarce labeled training examples~\cite{chen2018a,ye2020heterogeneous}.
Research literature on few-shot learning demonstrates great diversity.  Optimization-based methods~\cite{sun2020meta,maml,Sun_2019_CVPR,park2019meta,taml,omfsl,liu2020ensemble,yue2020interventional} and metric-based methods~\cite{prototypical,matchnet,feat,Zhang_2020_CVPR,Canfsl,gidaris2018dynamic,ye2019learning,zhang2020deepemdv2} are two main lines of efforts.
Optimization-based methods aims to design efficient learning paradigm that enables fast network adaptation given limited data~\cite{maml,Sun_2019_CVPR,park2019meta,taml,omfsl}.
Our work is more related to metric-based approaches, where a pre-trained backbone is used to encode data, and a distance metric, such as negative L2 distance~\cite{prototypical}, cosine similarity~\cite{matchnet} and DeepEMD~\cite{Zhang_2020_CVPR,zhang2020deepemdv2}, is used to measure data similarity and compute scores.
Chen~\etal~\cite{chen2018a} presents a baseline for few-shot classification that first pre-trains a backbone based on data from seen classes, and only finetunes the classifier for novel classes, which shares similarity with our decoupled training strategy.
Apart from image classification, few-shot learning has also been applied to dense  prediction tasks~\cite{zhang2019canet,liu2020crnet,chen2020compositional,pgnet,Liu_2021_CVPR} and object detection~\cite{yang2020context}.

\begin{figure*}[t]
	\centering
	\includegraphics[width=0.95\linewidth]{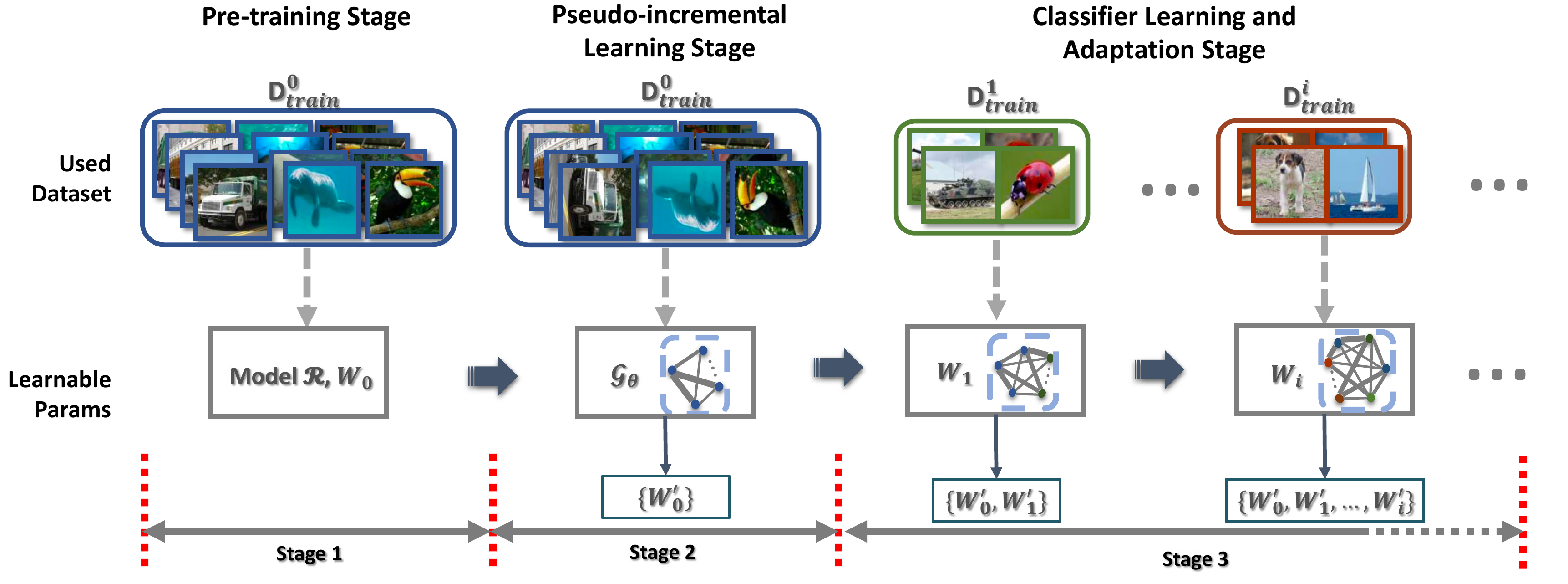}
	\caption{Our framework for few-shot incremental learning mainly includes three stages: (1) the feature pre-training stage to learn the backbone model $\mathcal{R}$ using the   training data in the base session $\mathcal{D}_{train}^0$, (2) the pseudo incremental learning stage that trains the graph model $\mathcal{G}_{\theta}$ by sampling  pseudo incremental tasks from $\mathcal{D}_{train}^0$, and (3) the classifier learning and adaptation stage using  few-shot training data  $\mathcal{D}_{train}^i$ in  new sessions }
	\label{fig:whole}
\vskip -1em
\end{figure*}

\textbf{Incremental Learning.} 
Incremental learning (IL) is an active machine learning task that aims to learn new knowledge continually without forgetting~\cite{2018efficient,dhar2019learning,chen2020mitigating,Liu2020AANets}. Recent works falls in two main streams, the multi-class incremental learning~\cite{castro2018end,iCaRL,hou2019learning,Liu_2020_CVPR,yu2020semantic,hu2021distilling} and the multi-task incremental learning~\cite{li2017learning,hu2018overcoming,riemer2018learning}.
Early approaches for IL use knowledge distillation~\cite{iCaRL,wu2019large} to transfer knowledge from the old model to a new model. iCaRL~\cite{iCaRL} learns a nearest-neighbor classifier with exemplars to preserve performance and combines distillation loss to avoid forgetting.
EEIL~\cite{castro2018end} introduces an end-to-end framework with cross-entropy loss and distillation loss for IL. 
LUCIR~\cite{hou2019learning} learns a unified classifier to solve the class imbalance problem between the base and new classes.
Liu~\etal~\cite{Liu_2020_CVPR} propose mnemonics training through bilevel optimizations in model-level and exemplar-level for tackling multi-class incremental learning.

\textbf{Few-Shot Class-Incremental Learning.} FSCIL~\cite{TOPIC,2020cognitively,ren19incfewshot,zhao2020few} is recently proposed with the goal of undertaking the CIL task with limited data in incremental sessions. It can also been seen as a few-shot learning task that can classify both novel and old classes at the same time.
Tao~\etal~\cite{TOPIC} propose a neural gas network to preserve the topology of the features in the base and new classes for the FSCIL task. Ren~\etal~\cite{ren19incfewshot} also undertake the few-shot incremental learning task but with a different setting.
Our work mainly follows the task definition proposed in~\cite{TOPIC} which is more closed to the setting in incremental learning literature.

\section{Problem Set-up}
FSCIL aims to design a machine learning algorithm that can continually learn  novel classes from only a few new training examples without forgetting knowledge about old classes.
Usually,  FSCIL has several learning sessions that come in sequence. Once the learning of model steps into the next session, the training dataset in previous learning sessions are no longer available, while the evaluation of the FSCIL algorithm in each session  involves  classes in all previous sessions and the current session.
To be specific,
let $\{ \mathcal{D}_{train}^0,\mathcal{D}_{train}^1,\cdots, \mathcal{D}_{train}^n \}$  denotes  the  training sets of different learning sessions, and the corresponding label space of dataset $\mathcal{D}_{train}^i$ is denoted by $\mathcal{C}^i$.
Different datasets have no overlapped classes, \ie  $\forall i,j $ and $i \neq j, \mathcal{C}^{i}\cap \mathcal{C}^{j} = \varnothing$. %
At the $i$th learning session, only $\mathcal{D}_{train}^i$ can be used for network training, and for evaluation, the test dataset $\mathcal{D}_{test}^{n}$ at session $i$ include test data from all previous and current classes, \ie, the label space of  $\mathcal{C}^0\cup \mathcal{C}^1 \cdots \cup \mathcal{C}^n$. 
Usually, the training set $\mathcal{D}_{train}^0$ in the first session  is a relatively large dataset where a sufficient amount of data is available for training, which is also called the base training set.
On the contrary, the datasets in all following sessions 
 have only a limited amount of data, and the dataset $\mathcal{D}_{train}^{i}$ on a specific session is often described as a $N-$way $K-$shot training set, where there are $N$ classes in the dataset, and each class has $K$ training images. 
For example, in the popular benchmark dataset CIFAR100, there are 60 classes in the base sessions, and each class has 500 training images, while in each incremental session, only 5 classes are available for training and each class only has 5 images.
FSCIL defines a harsh problem setting, where the severe data imbalance and scarcity problems will further exacerbate knowledge forgetting in incremental learning.

\section{Method}

In this section, we introduce our framework for few-shot incremental learning.
We first describe our decoupled training strategy of representations and classifiers in Section~\ref{section:decouple}. 
Then we present our proposed continually evolved classifier in Section~\ref{section:CEFC}.
To enable the learning of CEC, we design a pseudo incremental learning algorithm, which is described in Section~\ref{section:PIL}. 
The overview of the whole training pipeline is shown in Fig.~\ref{fig:whole}.

\subsection{Decoupling the Learning of Representations and Classifiers}
\label{section:decouple}
Our few-shot incremental learning framework mainly includes three training stages: the feature pre-training stage, the pseudo-incremental learning stage, and the classifier learning stage, as shown in Fig.~\ref{fig:whole}. The first two stages use data from the base sessions to learn the network backbone and the classifier adaptation module, and the classifier learning stage only learns the network classifier in each new incoming session.

\textbf{Feature pre-training stage.}
It is commonly evidenced in previous incremental learning literature that finetuning the network in new sessions can lead to significant knowledge forgetting of old classes. The data shortage problem in the few-shot incremental learning will further introduce the overfitting problem that exacerbates knowledge forgetting. 
To tackle this problem, we propose to decouple the learning of representations and classifiers to avoid the catastrophic forgetting issue at incremental stages.
Specifically, we first train a convolutional neural network in the standard manner  with the training dataset in the base session where abundant data are available for learning image representations, and we can then reuse the network backbone to encode image data in all sessions.
By freezing backbone parameters in new sessions, we can avoid knowledge forgetting and overfitting in the representations when learning the model on new sessions.

\textbf{Pseudo incremental learning stage.} Based on the pre-trained backbone model, we learn the classifier adaptation module to enable the function of the CEC, which is also based on the base dataset. The adaptation module is  frozen after training and is used to to update the classifiers learned on individual sessions.
We leave the detailed description of the classifier adaptation module and the training paradigm in Section~\ref{section:CEFC} and Section~\ref{section:PIL}. 

\textbf{Classifier learning stage.}  
Once the feature backbone and the graph models are learned in the base session, our model can be deployed for incremental learning.
We only need to learn a classifier upon the fixed backbone network with the dataset in new sessions, and then the learned classifiers in the current session and previous sessions are fed to the graph model for adaptation. Finally, the updated classifiers can be used for evaluation.

\subsection{Continually Evolved Classifier}
\label{section:CEFC}
As image categories come with groups in the incremental learning task, the classifiers learned on individual sessions may only provide discriminative decision boundaries between current classes. 
When all previous classes are involved for evaluation, the directly concatenated classifiers can not guarantee their discriminative ability and may fail to make correct decisions.
Therefore, to derive good decision boundaries over all classes, it is important to ensure the classifier learning incorporates the global context information of all individual tasks in previous sessions.
To achieve this goal, we propose a continually evolved classifier which includes a classifier adaptation module to update the classifier weights learned on each individual session based on the global context of previous sessions.
Let $\mathbf{W}_i \in \mathbb{R}^{N_i\times C} $ denotes the parameter matrix in the CNN classifier learned on session $i$, where each row vector $\vec{w}_i^{c}$ in $\mathbf{W}_i$ is the weights that correspond to a specific class $c$,  $N_i$ is the number of classes in session $i$ and $C$ is the number of feature channels. $\vec{w}_i^c$ can be seen as a prototype vector  for class $c$ where the values in different dimensions  essentially indicates the discriminability of different channels.
To refine the discriminability of the classifier, we can adjust the values in $\vec{w}_i^c$ by looking at $\vec{w}_i^i$ of all other classes.
To do so, we first collect the weight vectors of all other classes in the previous sessions:
\begin{equation}
\label{eq:proto}
\Tilde{\mathbf{W}_{I}}=\{
\vec{w}_0^1, \vec{w}_0^2,...,\vec{w}_i^{1},\vec{w}_i^{2},...,\vec{w}_I^{N_I}
\},
\end{equation}
where $I$ is the total number of sessions so far. 
Then, we use the Graph Attention Network (GAT)~\cite{velickovic2018graph} to model the relations between these prototype vectors and propagate context information, where all the weight vectors in $\Tilde{\mathbf{W}_{I}}$ can be regarded as the nodes in the graph model.
The Graph Attention Network has several desirable properties that make it an appropriate tool to encode context information: first, as the updating of graph node is based on attention mechanism, the context encoding is permutation invariant to the sequence of classes during incremental learning.
second, the GAT model allows a trained model to be extended to any number of classes, which means that the updating of classifiers at any sessions can share the same learned GAT. Since the nodes are fully connected in the GAT model, it has the similar structure with Transformer~\cite{vaswani2017attention} that also uses self-attention for information propagation.

To illustrate the context propagation process in the GAT, we take the updating of a node $j$ in the graph as an example. 
We first  compute  a relation coefficient $e_{jk}$ between the node $j$ and all nodes in the graph, such as $\vec{w}^j$ and  $\vec{w}^k$ :
\begin{equation}
e_{jk} = \langle \phi(\vec{w}^j),\theta(\vec{w}^k) \rangle,
\end{equation}
where, $\phi$ and $\theta$ are linear transformation functions that
project the original prototype representations to a new metric space. $\langle\cdot,\cdot\rangle$ is a similarity function that computes the inner product between two vectors.
Here we omit the session indexes in the subscript of  $\vec{w}^j$ and  $\vec{w}^k$  for clarity.
 We then normalize  all the coefficients  with the softmax function to get the final attention weights corresponding to the center node $j$:
\begin{equation}
a_{jk}=\text{softmax}(e_{jk})=\frac{\exp{(e_{jk}})}{\sum_{h=1}^{|\Tilde{\mathbf{W}} |}{\exp{(e_{jh})}}}.
\end{equation}
Based on the  normalized attention coefficients $a_{jk}$, we aggregate information from all the nodes in the graph based on  $a_{jk}$ and fuse it with the original node representation to obtain $\vec{w}^{j\prime}$:
\begin{equation}
\vec{w}^{j\prime}=\vec{w}^j + \big(\sum_{k=1}^{|\Tilde{\mathbf{W}_{I}}|}{a_{jk}\mathbf{U}\vec{w}^k}\big),
\end{equation}
where $\mathbf{U}$ is the weight matrix of a linear transformation.
We repeat the operations above to update the embeddings of all nodes in the graph, and finally we obtain the updated classifiers:
\begin{equation}
\Tilde{\mathbf{W}_{I}}'=\{
\vec{w}_0^{1\prime}, \vec{w}_0^{2\prime},...,\vec{w}_i^{1\prime},\vec{w}_i^{2\prime},...,\vec{w}_I^{N_I\prime}
\},
\end{equation}
In each incoming session, we use the adaptation module to update the classifiers learned in the current session and previous sessions, and then concatenate the updated classifiers to make predictions over all classes.
Many useful practices can be adopted to improve the knowledge propagation, such as multi-head attention~\cite{velickovic2018graph,vaswani2017attention}, layer normalization~\cite{vaswani2017attention}, and dropout~\cite{vaswani2017attention}. We also follow~\cite{feat} that incorporates the embedding of the network input into the graph to help the learning of context knowledge.

\begin{algorithm}[t]
\caption{Pseudo incremental learning.
$N_i$ is the number of classes in pseudo incremental classes;
$y_q$ and $\hat y_{q}$ indicates the ground truth label and the network predictions, respectively;
$\mathcal{L}(\cdot)$ is the cross-entropy loss function. 
}
\begin{algorithmic}[1]
\REQUIRE Base classes datasets $\mathcal{D}_{train}^0$, pre-trained model $\mathcal{R}$, a randomly initialized GAT model  $\mathcal{G}_{\theta}$.
\label{alg:PIL}
\ENSURE A trained GAT model $\mathcal{G}_{\theta}$.
\WHILE{not done} 

\STATE  $\{\mathcal{S}_{b}, \mathcal{Q}_{b}\} \leftarrow$ Sample the the support and query set for pseudo base classes from $\mathcal{D}_{train}^0$
\STATE $\mathbf{W}_{b} \leftarrow $ Learn FC layer upon $\mathcal{R}$ with $\mathcal{S}_{b}$
\STATE $\{\mathcal{S}_{i},\mathcal{Q}_{i}\} \leftarrow$ Sample the support and query set for pseudo incremental classes $\mathcal{D}_{train}^0$  
    \FOR{class $c$ \textbf{in} $N_{i}$}
        \STATE $\gamma \leftarrow$  Random select angle in
        $\{90^{\circ},180^{\circ},270^{\circ}\}$;
        \STATE $\{\mathcal{S}_{i}',\mathcal{Q}_{i}'\} \leftarrow$ Rotate $\{\mathcal{S}_{i},\mathcal{Q}_{i}\}$ from class $c$ with the selected angle $\gamma$;
    \ENDFOR
\STATE $\mathbf{W}_{i} \leftarrow $ Learn  FC layer upon $\mathcal{R}$ with  pseudo incremental support set $\mathcal{S}_{i}'$ after rotation
\STATE $\{\mathbf{W}_b',\mathbf{W}_{i}'\}\leftarrow$ Update classifier $\{\mathbf{W}_b,\mathbf{W}_{i}\}$ using $\mathcal{G}_{\theta}$
\STATE $\hat y_{q} \leftarrow$ Make predictions for $\{\mathcal{Q}_{b},\mathcal{Q}_{i}'\}$ using $[\mathcal{R},(\mathbf{W}_b',\mathbf{W}_{i}')]$ 
\STATE loss $\leftarrow$ Compute loss with $ \mathcal{L}(y_{q},\hat y_{q})$, 
\STATE Optimize $\mathcal{G}_{\theta}$ with SGD
\ENDWHILE
\end{algorithmic}
\end{algorithm}

\subsection{Pseudo Incremental Learning}
\label{section:PIL}
In order to enforce context encoding in the classifier adaptation module, it is important to learn the GAT under the incremental learning scenario.
However, in FSCIL, only data from a single session are available for training, and the amount of data in incremental sessions is always limited.
To overcome this problem, we design a pseudo incremental learning algorithm to train the adaptation module by episodically constructing pseudo incremental tasks based on the base dataset $\mathcal{D}_{train}^0$ to mimic the test scenario.  The pseudo code of the proposed algorithm is illustrated in Alg.~\ref{alg:PIL}.
Our algorithm takes inspirations from  meta-learning~\cite{matchnet}, where a small classification task is constructed to enable learning on the meta-level beyond a specific task.
We utilize data from the base dataset $\mathcal{D}_{train}^0$ to construct small incremental learning tasks for network training, where some sampled classes play the role of the base classes in incremental learning, while the other classes play the role of the incremental classes.
Specifically, both pseudo incremental classes and pseudo base classes have the support set and the query set, which are denoted by $(\mathcal{S}_{b},\mathcal{Q}_{b})$ and $(\mathcal{S}_{i},\mathcal{Q}_{i})$, respectively.
The support set is used to learn the classifier weights of different classes, and the query set is used to compute loss for optimization.
To be concrete, we first use the  support sets $(\mathcal{S}_{b}$ and $\mathcal{S}_{i})$ to learn two classifiers, $(\mathbf{W}_b'$ and $\mathbf{W}_{i}')$, for  pseudo base classes and pseudo incremental classes  respectively.
Then, the two classifiers are concatenated and fed into the adaptation module $\mathcal{G}_{\theta}$ for updating. 
We use the updated classifiers $(\mathbf{W}_b',\mathbf{W}_{i}')$ to make predictions for the query sets of pseudo base classes and pseudo incremental classes, \ie, $\mathcal{Q}_{b}$  and $\mathcal{Q}_{i}$, and compute the loss to optimize the adaptation module $\mathcal{G}_{\theta}$. 
We also finetune the last layer of the backbone with a small learning rate during PIL, which we find helpful.
In our experiment, we find that directly splitting the sampled base classes into two groups to train the adaptation module fails. A possible reason is that the backbone model  pre-trained on base classes can well separate these sampled classes already without context information. As a result, the training may simply bypass the adaptation module.
To handle this issue, we randomly rotate the data of the sampled pseudo incremental classes, $(\mathcal{S}_{i},\mathcal{Q}_{i})$,  with a large class-wise angle $\gamma$ to synthesis new classes, as we observe that rotating data with a large angle  can make the synthesized images lose parts of the semantics of their original classes, but demonstrate similar semantics among synthesized images. 
Once the adaptation module is learned, we can freeze the parameters in the adaptation module and deploy it in the new incremental sessions.

\begin{table*}[t]
\centering
\resizebox{1\textwidth}{!}{%
\begin{tabular}{lccccccccccccccc}
\toprule[1.2pt]
\multirow{2}{*}{\textbf{Method}} & \multirow{2}{*}{\textbf{Decoupled}}& \multirow{2}{*}{\textbf{AM}}& \multirow{2}{*}{\textbf{PIL}}& \multicolumn{11}{c}{\textbf{Acc. in each session (\%)} $\uparrow$} & \multirow{2}{*}{\textbf{PD} $\downarrow$} \\
\cline{5-15}
        &&& & \textbf{0}     & \textbf{1 }    & \textbf{2}     &\textbf{ 3 }    & \textbf{4}     & \textbf{5 }    & \textbf{6}     & \textbf{7}     & \textbf{8}     &\textbf{9}&\textbf{10}&             \\\hline\hline
\textbf{Linear} &  &&& 71.02 & 2.70  & 0.63 & 0.82 & 0.76 & 0.71 & 0.66 & 0.62 & 0.59 & 0.56 & 0.53 &70.49    \\ 
\textbf{Linear} & \checkmark  &&& 71.02  &7.22 &5.25 & 3.59 & 6.13 & 6.46 & 7.73 & 5.87 & 4.55 & 3.88 & 3.92 &67.10 \\
\Xhline{1pt}
\textbf{Cosine}  & &&& 73.32  &35.53 & 14.64 & 1.47 & 0.73 & 0.68 & 2.45 & 0.60 & 0.57 & 0.55 & 0.52&72.80 \\
\textbf{Cosine }& \checkmark && & 74.36  &48.50 & 43.40& 39.26 &37.29  & 33.69 & 33.36 & 32.49 &31.81  & 31.19 & 30.36 &44.00 \\
\Xhline{1pt}
\textbf{Linear+Data Init.} & \checkmark &  &&66.58&59.38 &54.29 &50.00&46.34 &43.18&40.43&38.00&35.85&34.55&32.76&33.83 \\
\textbf{L2+Data Init.} & \checkmark &  &&67.75&59.11 &56.05 &51.75 &51.39 &47.19&46.97&45.01&42.77&42.94&41.62&26.13 \\
\textbf{DeepEMD+Data Init.} & \checkmark &  &&75.35&70.69 &66.68 &62.34 &59.76 &56.54 & 54.61& 52.52&50.73& 49.20&47.60 &27.75   \\ 
\textbf{Cosine+Data Init.} & \checkmark &  &&75.52&70.95 &66.46 &61.20 &60.86 &56.88 & 55.40& 53.49&51.94& 50.93&49.31 &26.21    \\ 
\Xhline{1pt}
\textbf{Cosine+Data Init.} & \checkmark &\checkmark& &75.60   & 71.00 &66.89 & 61.81 &60.86  &56.81  &56.11  &53.59  &52.52  &50.59  &49.15 &  26.45\\
\textbf{Cosine+Data Init.} & \checkmark&\checkmark&\checkmark  &\textbf{75.85} &\textbf{71.94} &\textbf{68.50} &\textbf{63.50}  & \textbf{62.43} & \textbf{58.27} &\textbf{57.73 } &\textbf{55.81 } &\textbf{54.83}  &\textbf{53.52}  &\textbf{52.28}  &\textbf{23.57} \\
\Xhline{1.2pt}
\end{tabular}%
}
\caption{Ablation study on CUB200 to analyze the effectiveness of different components in our model. \textbf{AM} is the adaptation module, \textbf{PIL} is pseudo incremental learning, and \textbf{PD} denotes the performance dropping rate. }
\label{table:ablation}
\end{table*}

\section{Experiments}
 In this section, 
 we evaluate our proposed CEC on three popular few-shot incremental learning benchmark datasets, including CIFAR100~\cite{CIFAR100},  \emph{mini}ImageNet~\cite{imagenet} and Caltech-UCSD Birds-200-2011 (CUB200)~\cite{cub}.
We first present the experiment details and dataset statistics.
 Then we conduct comprehensive experiments to  validate the 
 the effectiveness of individual components in our design and study their characteristics.
 Finally, we compare our network with state-of-the-art methods on the benchmarks.

\subsection{Dataset}

\textbf{CIFAR100.} CIFAR100 is a classification dataset with 60,000  $32\times32$ RGB images from 100 classes. 
Each class contains 500 training images and 100 testing images. We follow the splits in~\cite{TOPIC}, where 60 classes and 40 classes are used as base classes and new classes, respectively. The 40 new classes are further divided into 8 new incremental sessions, and each new session is a 5-way 5-shot classification task.

\textbf{\emph{mini}ImageNet.} \emph{mini}ImageNet  contains 100 classes with 600 images in each class, which are built upon the ImageNet dataset~\cite{imagenet}.
The image size of \emph{mini}ImageNet is $84\times84$ and we follow~\cite{TOPIC} to split the 100 classes into 60 base classes and 40 incremental classes. The 40 new classes are further divided equally into 8 sessions with 5 classes in each session, and each class has 5 training images in the incremental sessions.

\textbf{Caltech-UCSD Birds-200-2011.} CUB200~\cite{cub} was originally proposed for fine-grained image classification. It contains 11,788 images from 200 classes.
We follow the splits in~\cite{TOPIC} that 200 classes are divided into 100 base classes and 100 new classes, respectively. The 100 new classes are further divided into 10 new sessions where each session is a  10-way 5-shot task.
The images size in CUB200 is $224\times224$.

\begin{figure}[t]
\vskip -1em
	\centering
	\includegraphics[width=1\linewidth]{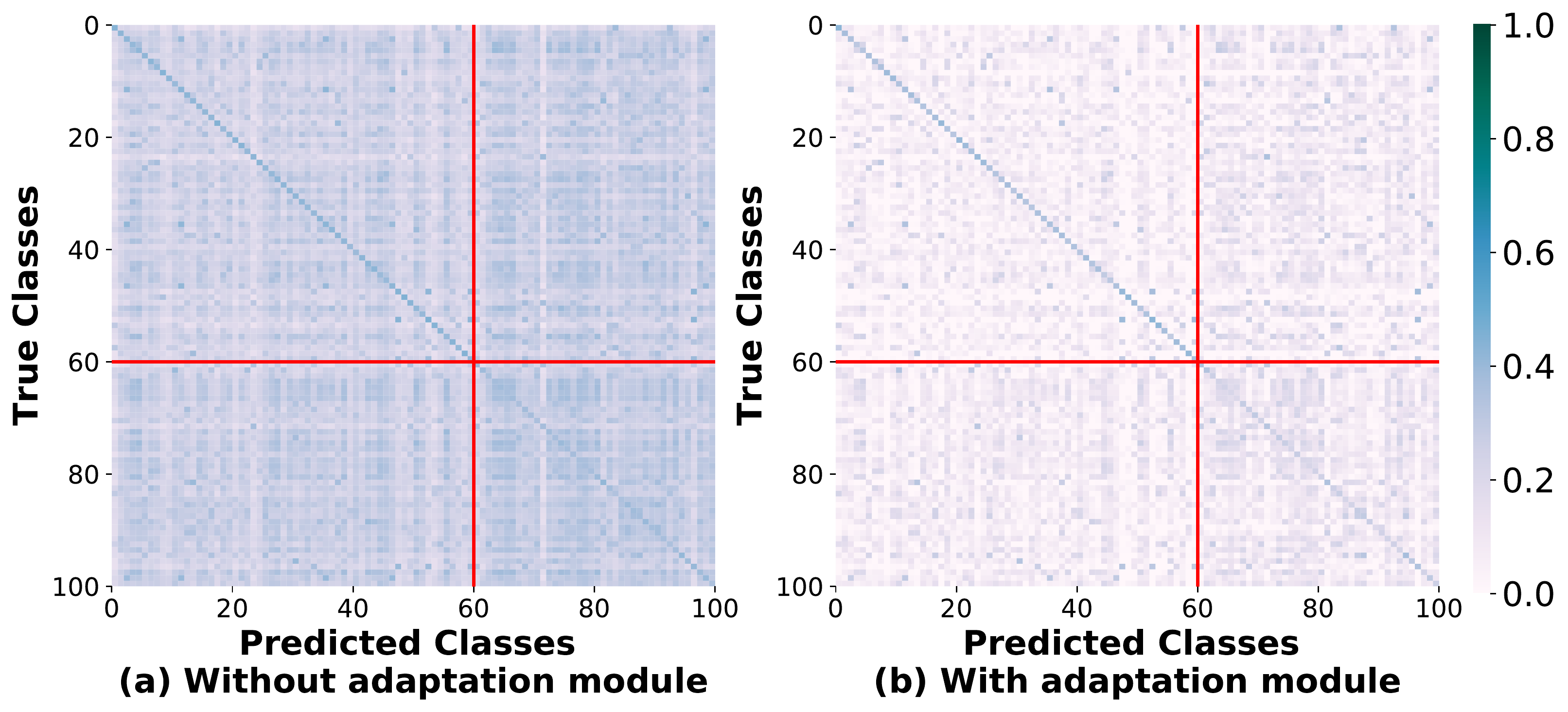}
	\caption{Confusion matrices with and without adaptation module on CIFAR100. 
	We use red lines to separate regions of base classes and incremental classes. Our adaptation module effectively improves the network prediction, which results in a less scattered confusion matrix.}
	\label{fig:graph_fig}
\vskip -1em
\end{figure}

\subsection{Implementation Details}

Following~\cite{TOPIC}, we employ ResNet20~\cite{resnet} as the backbone for experiments on CIFAR100 and ResNet18~\cite{resnet} for experiments on miniImageNet and CUB200. 
Our network is built with PyTorch library, and we use SGD with momentum for optimization.
At the pseudo incremental learning stage, we random choose the angle $\gamma$ from $\{90^{\circ},180^{\circ},270^{\circ}\}$ to synthesize new classes.
We train the graph model $\mathcal{G}_{\theta}$ for 5000 iterations with the learning rate of 0.0002. The learning rate is decayed by 0.5 every 1000 iteration. Random crop, random scale, and random horizontal flip are used for data augmentation at training time.

\textbf{Evaluation Protocol.}
We evaluate the model after each session with the test set $\mathcal{D}_{test}^{i}$ and report the Top 1 accuracy.
We also define a performance dropping rate (\textbf{PD}) that measures the absolute accuracy drops in the last session w.r.t. the accuracy in the first session,
\ie, $\text{PD}=\mathcal{A}_0-\mathcal{A}_{N}$, where $\mathcal{A}_0$ is the classification accuracy in the base session  and $\mathcal{A}_{N}$ is the accuracy in the last session.

\begin{figure}[t]
	\centering
	\includegraphics[width=1\linewidth]{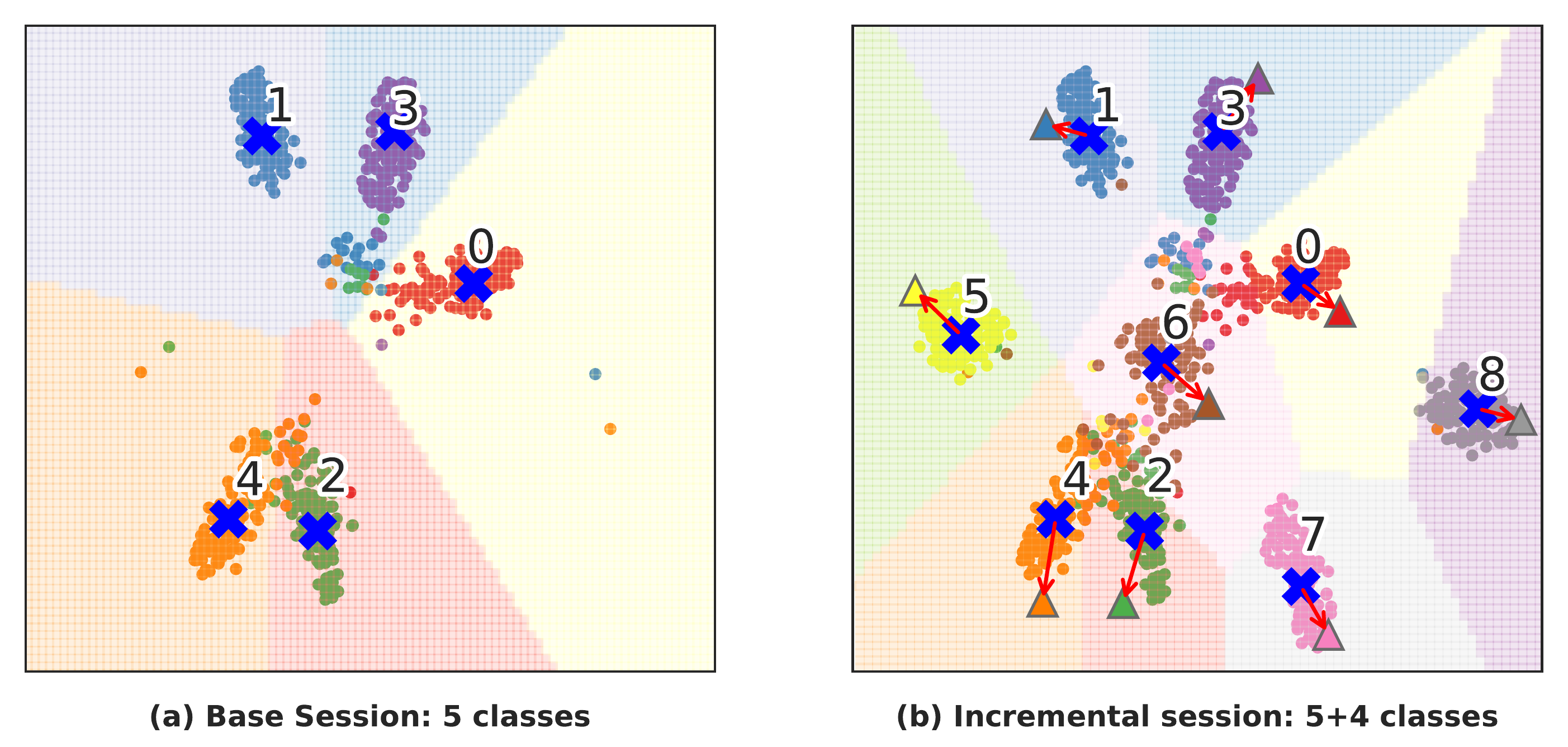}
	\caption{
	t-SNE~\cite{maaten2008visualizing} visualization of data embeddings and classifier weights before and after the adaptation module.
	Dots with different colors represent data points from different classes. The blue crosses indicate the classifier weights before adaptation. Triangles indicate the weights after adaptation. Red arrows show the changes in weights caused by the adaptation module. Our adaptation module moves the classifier weights away from the confusion area and generates better decision boundaries.}
	\label{fig:tsne_fig}
\vskip -1em
\end{figure}

\begin{figure}[t]
	\centering
	\includegraphics[width=0.8\linewidth]{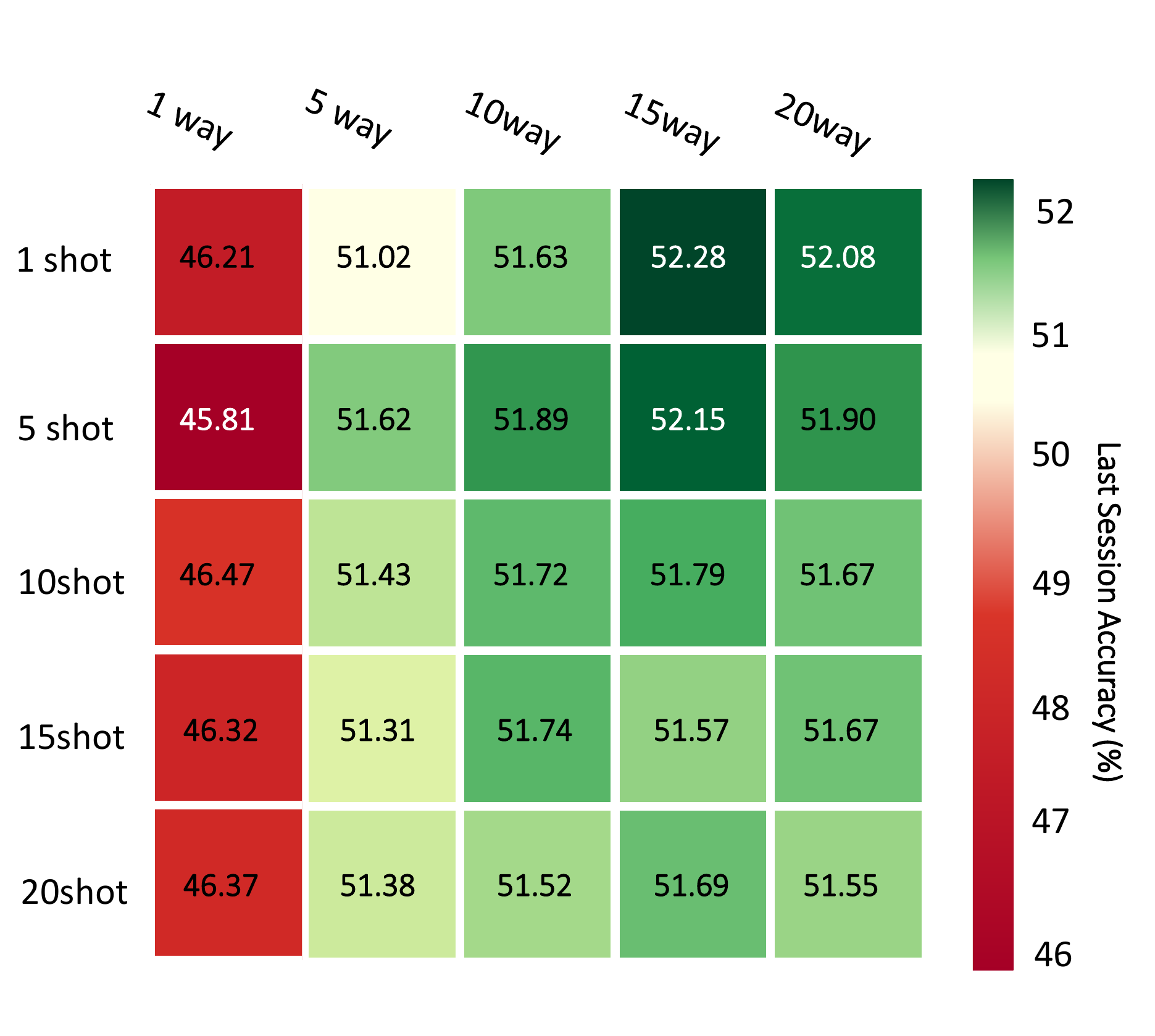}
	\caption{Comparison of different ways and shots for pseudo incremental learning. We report the accuracy of the last incremental session on the CUB200~\cite{cub} dataset for comparison. Large ways and low shots are preferred for learning of the adaptation module.}
	\label{fig:wayshot_fig}
\vskip -1em
\end{figure}

\subsection{Analysis}
In this part, we implement various experiments to evaluate the effectiveness of our algorithm and study the characteristics of different components. For analysis, we mainly report the results on the CUB200 dataset and leave other datasets in Section~\ref{section:sota} and our supplementary material.

\textbf{Ablation study.} In the beginning, we conduct an ablative analysis  on the CUB200 dataset to observe the effectiveness of the different components in our model. 
We first consider four kinds of classifiers, including the vanilla \textbf{linear} classifier in the CNNs, the \textbf{cosine classifier}~\cite{matchnet}, the \textbf{L2} classifier~\cite{prototypical}, and the \textbf{DeepEMD} classifier~\cite{Zhang_2020_CVPR}, where their main difference is the metric to compute class scores given the prototypes of each class.
In new incremental sessions, the classifier is learned  with a learning rate of 0.1 for 100 epochs. We also try using the data embeddings to parameterize the classifier weights where the weight vector of each class is initialized by the average data embeddings in the training set, which is denoted by \textbf{Data Init}.
We gradually involve  our designs to observe their influence on performance, including decoupled training scheme (\textbf{Decoupled}), the adaptation module (\textbf{AM}), and the pseudo incremental learning paradigm (\textbf{PIL}). When our adaptation module is not trained with pseudo incremental learning, we adopt the meta-learning~\cite{matchnet} to learn the parameters in the graph.
 The result is shown in Table~\ref{table:ablation}.
 For both cosine classifier and the linear classifier, decoupling the representation learning and the classifier learning is useful for avoiding the catastrophic forgetting issue, which can  decrease the performance dropping rate by 28.81\% and 3.39\%, respectively.
 Using the data embeddings to initialize the classifier weights is beneficial to all classifiers. 
 When both the decoupled training strategy and data initialization are adopted, all four classifiers can achieve good performance, and cosine classifier performs the best.
Without further specification, we use the cosine classifier in rest experiments.
 Using meta-learning to learn the adaptation module fails to improve the performance. When the adaptation module is learned by our proposed PIL, it can boost the performance over all sessions by up to 3.13\% and can decrease performance dropping rate by 2.87\%.

\begin{figure}[t]
\centering
\centering
	\includegraphics[width=1\linewidth]{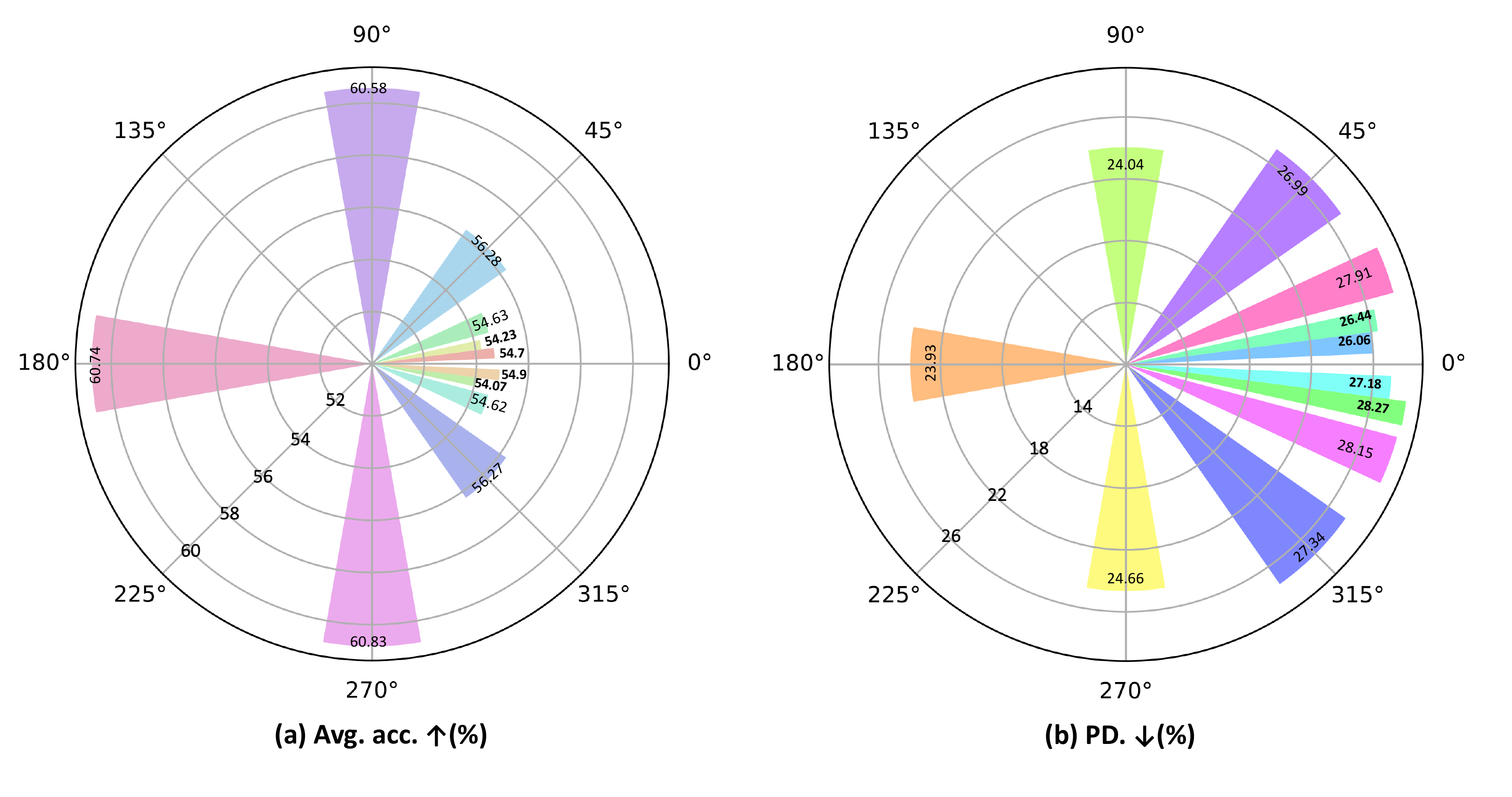}
	\caption{Comparison of different rotation degrees for pseudo incremental learning. Our tested degrees include $180^{\circ}$, $\pm 90^{\circ}$, $\pm 45^{\circ}$, $\pm 20^{\circ}$, $\pm 10^{\circ}$ and $\pm 5^{\circ}$.
	We report the average accuracy of all the sessions and the performance dropping (PD) on the CUB200 dataset for comparison. Large rotation degrees, such as $180^{\circ}$ and $\pm 90^{\circ}$, are more helpful for class synthesis. }
	\label{fig:rotation}

\end{figure}

\begin{figure*}[t]
	\centering
	\includegraphics[width=0.95\linewidth]{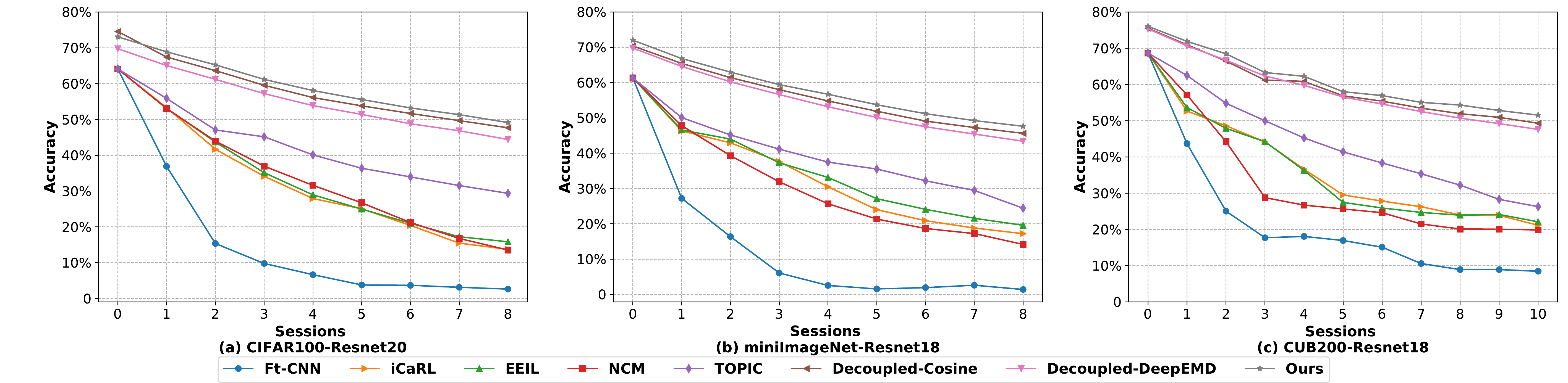}
	\caption{
	Comparison with the state-of-the-art on three benchmarks: (a) CIFAR100 (b) miniImageNet and (c) CUB200. Our method outperforms previous works with significant performance advantages. Please refer to  Table~\ref{table:sota} and our supplementary material for detailed numbers.}
	\label{fig:sota}
\end{figure*}

\begin{table*}[t]
\centering
\resizebox{0.95\textwidth}{!}{
\begin{tabular}{lccccccccccccc}
\toprule[1pt]
\multirow{2}{*}{Method} & \multicolumn{11}{c}{Acc. in each session (\%) $\uparrow$} & \multirow{2}{*}{PD $\downarrow$} & \multirow{2}{*}{\shortstack[c]{our relative\\improvement}}\\ \cline{2-12}
         & 0     & 1     & 2     & 3     & 4     & 5     & 6     & 7     & 8     &9&10&     &        \\ \Xhline{1pt}
Ft-CNN & 68.68  & 43.7   & 25.05  & 17.72  & 18.08 & 16.95  & 15.1   & 10.6  & 8.93   & 8.93   & 8.47   & 60.21   &\textbf{+36.64}                 \\
iCaRL*~\cite{iCaRL} & 68.68  & 52.65  & 48.61  & 44.16  & 36.62 & 29.52  & 27.83  & 26.26 & 24.01  & 23.89  & 21.16  & 47.52&  \textbf{+23.95}                    \\
EEIL*~\cite{castro2018end}  & 68.68  & 53.63  & 47.91  & 44.2   & 36.3  & 27.46  & 25.93  & 24.7  & 23.95  & 24.13  & 22.11  & 46.57    &\textbf{+23.00}                \\
NCM*~\cite{hou2019learning}   & 68.68  & 57.12  & 44.21  & 28.78  & 26.71 & 25.66  & 24.62  & 21.52 & 20.12  & 20.06  & 19.87  & 48.81    &\textbf{+25.24}                \\
TOPIC~\cite{TOPIC}  & 68.68  & 62.49  & 54.81  & 49.99  & 45.25 & 41.4   & 38.35  & 35.36 & 32.22  & 28.31  & 26.28  & 42.40  &\textbf{+18.83}            \\
\hline
Decoupled-Cosine~\cite{matchnet}${}^{\ddag}$ &75.52&70.95 &66.46 &61.20 &60.86 &56.88 & 55.40& 53.49&51.94& 50.93&49.31 &26.21&\textbf{+2.64} \\
Decoupled-DeepEMD~\cite{Zhang_2020_CVPR}${}^{\ddag}$ &75.35&70.69 &66.68 &62.34 &59.76 &56.54 & 54.61& 52.52&50.73& 49.20&47.60 &27.75&\textbf{+4.18}   \\
\hline
\textbf{CEC (Ours)} &\textbf{75.85} & \textbf{71.94} & \textbf{68.50} & \textbf{63.5} & \textbf{62.43} & \textbf{58.27} & \textbf{57.73} &\textbf{55.81} &  \textbf{54.83} & \textbf{53.52} &  \textbf{52.28} &\textbf{23.57}  &     \\ \Xhline{1pt}
\multicolumn{10}{l}{ ${}^{\ddag}$ Our implementation.}
\end{tabular}}
\caption{Comparison with the state-of-the-art on CUB200 dataset.  * indicates results copied from TOPIC~\cite{TOPIC}.  Please refer to our supplementary material for the detailed results on other datasets. }
\label{table:sota}

\end{table*}

\textbf{Confusion Matrix.} To further observe the behavior in the adaptation module, we plot the confusion matrix generated by the models with and without our adaptation module in Fig.~\ref{fig:graph_fig}. 
As we can see, the classifier without adaptation generates a confusing matrix, particularly for the incremental classes (the prediction distribution is more scattered and thus darker). In contrast, our adaptation module can effectively improve the predictions where the values  more lie in the diagonal that indicates the ground truth.

\textbf{Visualization of adaptation.} We plot the data embeddings and classifier weights in low-dimension space with t-SNE~\cite{maaten2008visualizing} in Fig.~\ref{fig:tsne_fig}. 
We randomly choose five classes from the CIFAR100 dataset as the base classes, and we add four new classes as the incremental classes.
 As can be seen, the adaptation module moves the classifier weights away from the confusion area to generate better decision boundaries when new classes are involved.

\textbf{Analysis of pseudo incremental learning.}  We next investigate the configurations in the pseudo incremental learning scheme. In particular, we fix the query number as 10 and  analyze the influence of ways, shots and the rotation angles during pseudo incremental learning.
We set the same ways, shots and queries for pseudo base classes and pseudo incremental classes. 
The comparison is shown in  Fig.~\ref{fig:wayshot_fig}.  We choose the number of ways from $\{1, 5, 10, 15, 20\}$ and the number of shots from $\{1, 5, 10, 15, 20\}$.
We find that a relatively larger way and a smaller shot are better, and the optimal result is obtained when the way is 15 and the shot is 1. 

We then fix the way and shot, and investigate the rotation degrees for classes synthesis in PIL. 
We choose different rotation degrees for comparison and  present their results in Fig.~\ref{fig:rotation}. Our tested degrees include $180^{\circ}$, $\pm 90^{\circ}$, $\pm 45^{\circ}$, $\pm 20^{\circ}$, $\pm 10^{\circ}$ and $\pm 5^{\circ}$.
As we can see, large angles, such as $180^{\circ}$, $90^{\circ}$ and $-90^{\circ}$ ($270^{\circ}$) are more effective for class synthesis and generate higher average accuracy and lower performance dropping rate. When the rotation degree is small, the synthesized classes may be confused with the original classes and thus generate poor results. When the three large degrees, \ie, \{$180^{\circ}$, $90^{\circ}$, $-90^{\circ}$($270^{\circ}$) \} are randomly selected for training, it generates the best result with the highest average accuracy of $61.33\%$ and lowest dropping rate of $23.57\%$.

\subsection{Comparison with the State-of-the-Art Methods}
\label{section:sota}
Finally, we compare our performance with the state-of-the-art results on three benchmarks: CIFAR100, miniImagenet, and CUB200. 
We show the results in Fig.~\ref{fig:sota} and the detailed numbers for CUB200 in Table~\ref{table:sota} (Please refer to our supplementary material for results on other datasets).
Our model has the highest average accuracy over all sessions and the lowest performance dropping rate.
 Particularly, our PD outperforms the state-of-the-art results by 10.80\% on CIFAR100, 12.52\% on \emph{mini}ImageNet and 18.83\% on CUB200.

\section{Conclusion}

In this paper, we solve the few-shot incremental learning problems from two aspects. We first  
adopt a decoupled learning strategy to separate the learning of representations and classifiers, which effectively avoid knowledge forgetting in the backbone. Then, we propose a continually evolved classifier for few-shot incremental learning, which employs an adaptation module to  update the classifier weights based on a global context of all sessions. To enable the learning of the adaptation module, we propose a  pseudo incremental learning paradigm.
Experiments on three datasets show that our method significantly outperforms the baselines and the state-of-the-art approaches.

\section*{Acknowledgement}

This work was supported by Alibaba Group through Alibaba Innovative Research (AIR) Program and Alibaba-NTU Singaproe Joint Research Institute (JRI), Nanyang Technological University, Singapore. This research is also supported by the National Research Foundation, Singapore under its AI Singapore Programme (AISG Award No: AISG-RP-2018-003), and the MOE Tier-1 research grants: RG28/18 (S), RG22/19 (S) and RG95/20.

{\small
\bibliographystyle{ieee_fullname}
\bibliography{egbib}
}
\clearpage
\onecolumn
\appendix
\begin{center}
\textbf{\large Supplementary Material}
\end{center}
\section{Introduction}
In our supplementary material, we present more details about the experiments in our paper.
\section{Detailed Result}
In Section~\ref{section:sota} Fig.~\ref{fig:sota}, we have provided the comparison with the state-of-the-art methods in the form of line charts. Here, we present the detailed numbers in Table~\ref{table:sotafull}. The results show that our method  significantly outperforms the baselines and achieves new state-of-the-art performance on all the three datasets.

\begin{table*}[hbt]
\small
\centering

\begin{subtable}{\textwidth}
\centering
\resizebox{0.95\textwidth}{!}{
\small

\begin{tabular}{lccccccccccc}
\toprule[1pt]
\multirow{2}{*}{Method} & \multicolumn{9}{c}{Acc. in each session (\%) $\uparrow$} & \multirow{2}{*}{PD $\downarrow$} & \multirow{2}{*}{\shortstack[c]{our relative\\improvement}} \\ \cline{2-10}
         & 0     & 1     & 2     & 3     & 4     & 5     & 6     & 7     & 8     &    &         \\ \Xhline{1pt}
Ft-CNN   & 64.1  & 36.91 & 15.37 & 9.8   & 6.67  & 3.8   & 3.7   & 3.14  & 2.65  & 61.45 &  \textbf{+37.52} \\
iCaRL*~\cite{iCaRL}   & 64.1  & 53.28 & 41.69 & 34.13 & 27.93 & 25.06 & 20.41 & 15.48 & 13.73 & 50.37  &  \textbf{+26.44}\\
EEIL*~\cite{castro2018end}    & 64.1  & 53.11 & 43.71 & 35.15 & 28.96 & 24.98 & 21.01 & 17.26 & 15.85 & 48.25  &  \textbf{+24.32}\\
NCM*~\cite{hou2019learning}     & 64.1  & 53.05 & 43.96 & 36.97 & 31.61 & 26.73 & 21.23 & 16.78 & 13.54 & 50.56 &  \textbf{+26.63}\\
TOPIC~\cite{TOPIC}    & 64.1  & 55.88 & 47.07 & 45.16 & 40.11 & 36.38 & 33.96 & 31.55 & 29.37 & 34.73 & \textbf{+10.80}\\ \hline
Decoupled-Cosine~\cite{matchnet}${}^{\ddag}$ & 74.55 & 67.43 & 63.63 & 59.55 & 56.11 & 53.80 & 51.68 & 49.67 & 47.68& 26.87 & \textbf{+2.94}\\
Decoupled-DeepEMD~\cite{Zhang_2020_CVPR}${}^{\ddag}$ & 69.75 &65.06 & 61.2 & 57.21 & 53.88 & 51.40 & 48.80 & 46.84 & 44.41 & 25.34  & \textbf{+1.41}  \\
\hline
\textbf{CEC (Ours)} & \textbf{73.07} & \textbf{68.88} & \textbf{65.26} & \textbf{61.19} & \textbf{58.09} & \textbf{55.57} & \textbf{53.22} & \textbf{51.34} & \textbf{49.14} & \textbf{23.93}  &     \\ \Xhline{1pt}
\end{tabular}}
\caption{CIFAR100 results using 5-way 5-shot FSCIL setting }
\end{subtable}
\label{table:cifar100}
\vskip 1em
\begin{subtable}{\textwidth}
\centering
\resizebox{0.95\textwidth}{!}{
\begin{tabular}{lcccccccccccc}
\toprule[1pt]
\multirow{2}{*}{Method} & \multicolumn{9}{c}{Acc. in each session (\%) $\uparrow$} & \multirow{2}{*}{PD $\downarrow$} & \multirow{2}{*}{\shortstack[c]{our relative\\improvement}} \\ \cline{2-10}
         & 0     & 1     & 2     & 3     & 4     & 5     & 6     & 7     & 8     &   &          \\ \Xhline{1pt}
Ft-CNN & 61.31 & 27.22 & 16.37 & 6.08  & 2.54  & 1.56  & 1.93  & 2.6   & 1.4   & 59.91  &  \textbf{+35.54}  \\
iCaRL*~\cite{iCaRL} & 61.31 & 46.32 & 42.94 & 37.63 & 30.49 & 24    & 20.89 & 18.8  & 17.21 & 44.10   &  \textbf{+19.73} \\
EEIL*~\cite{castro2018end}  & 61.31 & 46.58 & 44    & 37.29 & 33.14 & 27.12 & 24.1  & 21.57 & 19.58 & 41.73 &  \textbf{+17.36}\\
NCM*~\cite{hou2019learning}   & 61.31 & 47.8  & 39.31 & 31.91 & 25.68 & 21.35 & 18.67 & 17.24 & 14.17 & 47.14  &  \textbf{+22.77}\\
TOPIC~\cite{TOPIC}  & 61.31 & 50.09 & 45.17 & 41.16 & 37.48 & 35.52 & 32.19 & 29.46 & 24.42 & 36.89  & \textbf{+12.52}\\\hline
Decoupled-Cosine~\cite{matchnet}${}^{\ddag}$ & 70.37 & 65.45 & 61.41 & 58.00 & 54.81 & 51.89 & 49.10 & 47.27 & 45.63 & 24.74 & \textbf{+0.37}\\
Decoupled-DeepEMD~\cite{Zhang_2020_CVPR}${}^{\ddag}$ & 69.77 & 64.59 & 60.21 & 56.63 & 53.16 & 50.13 & 47.49 & 45.42 & 43.41 & 26.36  & \textbf{+1.99}  \\
\hline
\textbf{CEC (Ours)} &\textbf{72.00} & \textbf{66.83} & \textbf{62.97} & \textbf{59.43} & \textbf{56.70} & \textbf{53.73} &\textbf{51.19} &\textbf{49.24} & \textbf{47.63} & \textbf{24.37} &    \\ \Xhline{1pt}
\end{tabular}}
\caption{miniImageNet results using 5-way 5-shot FSCIL setting}
\end{subtable}
\label{table:miniImagenet}
\vskip 1em

\begin{subtable}{\textwidth}
\centering
\resizebox{0.95\textwidth}{!}{
\begin{tabular}{lccccccccccccc}
\toprule[1pt]
\multirow{2}{*}{Method} & \multicolumn{11}{c}{Acc. in each session (\%) $\uparrow$} & \multirow{2}{*}{PD $\downarrow$} & \multirow{2}{*}{\shortstack[c]{our relative\\improvement}}\\ \cline{2-12}
         & 0     & 1     & 2     & 3     & 4     & 5     & 6     & 7     & 8     &9&10&     &        \\ \Xhline{1pt}
Ft-CNN & 68.68  & 43.7   & 25.05  & 17.72  & 18.08 & 16.95  & 15.1   & 10.6  & 8.93   & 8.93   & 8.47   & 60.21   &\textbf{+36.64}                 \\
iCaRL*~\cite{iCaRL} & 68.68  & 52.65  & 48.61  & 44.16  & 36.62 & 29.52  & 27.83  & 26.26 & 24.01  & 23.89  & 21.16  & 47.52&  \textbf{+23.95}                    \\
EEIL*~\cite{castro2018end}  & 68.68  & 53.63  & 47.91  & 44.2   & 36.3  & 27.46  & 25.93  & 24.7  & 23.95  & 24.13  & 22.11  & 46.57    &\textbf{+23.00}                \\
NCM*~\cite{hou2019learning}   & 68.68  & 57.12  & 44.21  & 28.78  & 26.71 & 25.66  & 24.62  & 21.52 & 20.12  & 20.06  & 19.87  & 48.81    &\textbf{+25.24}                \\
TOPIC~\cite{TOPIC}  & 68.68  & 62.49  & 54.81  & 49.99  & 45.25 & 41.4   & 38.35  & 35.36 & 32.22  & 28.31  & 26.28  & 42.40  &\textbf{+18.83}            \\
\hline
Decoupled-Cosine~\cite{matchnet}${}^{\ddag}$ &75.52&70.95 &66.46 &61.20 &60.86 &56.88 & 55.40& 53.49&51.94& 50.93&49.31 &26.21&\textbf{+2.64} \\
Decoupled-DeepEMD~\cite{Zhang_2020_CVPR}${}^{\ddag}$ &75.35&70.69 &66.68 &62.34 &59.76 &56.54 & 54.61& 52.52&50.73& 49.20&47.60 &27.75&\textbf{+4.18}   \\
\hline
\textbf{CEC (Ours)} &\textbf{75.85} & \textbf{71.94} & \textbf{68.50} & \textbf{63.5} & \textbf{62.43} & \textbf{58.27} & \textbf{57.73} &\textbf{55.81} &  \textbf{54.83} & \textbf{53.52} &  \textbf{52.28} &\textbf{23.57}  &     \\ \Xhline{1pt}
\multicolumn{10}{l}{ ${}^{\ddag}$ Our implementation.}
\end{tabular}}
\caption{CUB200 results using 10-way 5-shot FSCIL setting}
\end{subtable}
\caption{Comparison with the state-of-the-art on (a) CIFAR100, (b) miniImagenet and (c) CUB200 datasets.  * indicates results copied from TOPIC~\cite{TOPIC}. ${}^{\ddag}$ indicates our implementation for the method under FSCIL setting. Our method outperforms the state-of-the-art results with large advantages on three benchmarks. }
\label{table:sotafull}
\end{table*}

\clearpage
\end{document}